\let\accentvec\vec
\documentclass[runningheads,a4paper]{llncs}

\let\vec\accentvec
\usepackage{amssymb}
\usepackage{amsmath}
\usepackage{bbm}
\usepackage{xfrac}
\usepackage{graphicx}
\usepackage{cite}
\usepackage{multirow}
\usepackage{booktabs}
\usepackage{tabularx}
\usepackage{array}
\usepackage{epstopdf}
\newcolumntype{+}{>{\global \let \currentrowstyle \relax }}
\newcolumntype{^}{>{\currentrowstyle }}

\usepackage{url}
\urldef{\mailsa}\path| *@*|

\newcolumntype{Y}{>{\centering\arraybackslash}X}
\begin{document}
\mainmatter 
\title{Robust Image Descriptors for Real-Time Inter-Examination Retargeting in Gastrointestinal Endoscopy}
\titlerunning{Robust Image Descriptors for Real-Time Inter-Examination Retargeting}
%
%
\author{Menglong Ye$^{1}$, Edward Johns$^{2}$, Benjamin Walter$^{3}$, Alexander Meining$^{3}$ and Guang-Zhong Yang$^{1}$
}
\authorrunning{M. Ye et al.}

\institute{$^{1}$The Hamlyn Centre for Robotic Surgery, Imperial College London, UK\\
$^{2}$Dyson Robotics Laboratory, Imperial College London, UK\\
$^{3}$Centre of Internal Medicine, Ulm University, Germany\\
\url{menglong.ye11@imperial.ac.uk}
}
%
%
\maketitle
\begin{abstract}
For early diagnosis of malignancies in the gastrointestinal tract, surveillance endoscopy is increasingly used to monitor abnormal tissue changes in serial examinations of the same patient. Despite successes with optical biopsy for \emph{in vivo} and \emph{in situ} tissue characterisation, biopsy retargeting for serial examinations is challenging because tissue may change in appearance between examinations. In this paper, we propose an inter-examination retargeting framework for optical biopsy, based on an image descriptor designed for matching between endoscopic scenes over significant time intervals. Each scene is described by a hierarchy of regional intensity comparisons at various scales, offering tolerance to long-term change in tissue appearance whilst remaining discriminative. Binary coding is then used to compress the descriptor via a novel random forests approach, providing fast comparisons in Hamming space and real-time retargeting. Extensive validation conducted on 13 \emph{in vivo} gastrointestinal videos, collected from six patients, show that our approach outperforms state-of-the-art methods.
\end{abstract}
\section{Introduction}
In gastrointestinal (GI) endoscopy, serial surveillance examinations are increasingly used to monitor recurrence of abnormalities, and detect malignancies in the GI tract in time for curative therapy. In addition to direct visualisation of the mucosa, serial endoscopic examinations involve the procurement of histological samples from suspicious regions, for diagnosis and assessment of pathologies. Recent advances in imaging modalities such as confocal laser endomicroscopy and narrow band imaging (NBI), allow for \emph{in vivo} and \emph{in situ} tissue characterisation with optical biopsy. Despite the advantages of optical biopsy, the required retargeting of biopsied locations, for tissue monitoring, during intra- or inter-examination of the same patient is challenging. 

For intra-examination, retargeting techniques using local image features have been proposed, which include feature matching \cite{Atasoy2009}, geometric transformations \cite{Allain2012}, tracking \cite{Ye2013, Ye2014}, and mapping \cite{Mountney2009}. However, when applied over successive examinations, these often fail due to the long-term variation in appearance of tissue surface, which causes difficulty in detecting the same local features. For inter-examination, endoscopic video manifolds (EVM) \cite{Atasoy2012} was proposed, with retargeting achieved by projecting query images into manifold space using locality preserving projections. In \cite{Vemuri2013}, an external positioning sensor was used for retargeting, but requiring manual trajectory registration which interferes with the clinical workflow, increasing the complexity and duration of the procedure. 

\begin{figure*}[t]
\centering
\includegraphics[width=\linewidth]{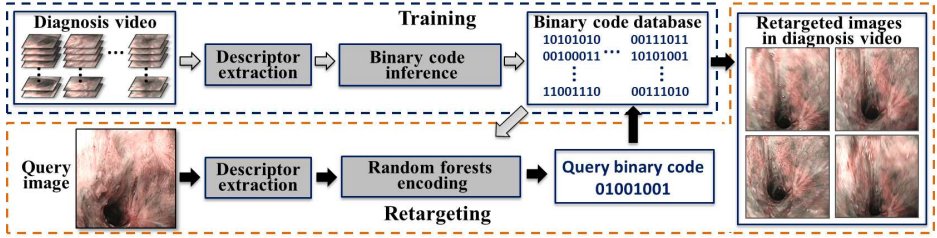}
\caption{A framework overview. Grey arrows represent the training phase using diagnosis video while black arrows represent the querying phase in the surveillance examination.}
\label{fig:workflow}
\end{figure*}
In this work, we propose an inter-examination retargeting framework (see Fig.\ref{fig:workflow}) for optical biopsy. This enables recognition of biopsied locations in the surveillance (second) examination, based on targets defined in the diagnosis (first) examination, whilst not interfering with the clinical workflow. Rather than relying on feature detection, a global image descriptor is designed based on regional image comparisons computed at multiple scales. At the higher scale, this offers robustness to small variations in tissue appearance across examinations, whilst at the lower scale, this offers discrimination in matching those tissue regions which have not changed. Inspired by \cite{Lin2015}, efficient descriptor matching is achieved by compression into binary codes, with a novel mapping function based on random forests, allowing for fast encoding of a query image and hence real-time retargeting. Validation was performed on $13$ \emph{in vivo} GI videos, obtained from successive endoscopies of the same patient, with 6 patients in total. Extensive comparisons to state-of-the-art methods have been conducted to demonstrate the practical clinical value of our approach.
 
\section{Methods}
\subsection{A Global Image Descriptor for Endoscopic Scenes}
Visual scene recognition is often addressed using keypoint-based methods such as SIFT \cite{Lowe2004}, typically made scalable with Bag-of-Words (BOW) \cite{Philbin2007}. However, these approaches rely on consistent detection of the same keypoint on different observations of the same scene, which is often not possible when applied to endoscopic scenes undergoing long-term appearance changes of the tissue surface.

\begin{figure*}[t]
\centering
\includegraphics[width=\linewidth]{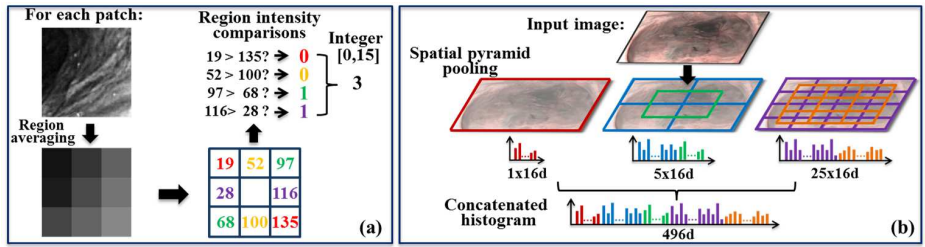}
\caption{(a) Obtaining an integer from one location; (b) Creating the global image descriptor from all locations using spatial pyramid pooling.}
\label{fig:blbp}
\end{figure*}
In recent years, the use of local binary patterns (LBP) \cite{Wu2011} has proved popular for recognition due to its fast computational speed, and robustness to image noise and illumination variation. Here, pairs of pixels within an image patch are compared in intensity to create a sequence of binary numbers. We propose a novel, symmetric version of LBP which performs 4 diagonal comparisons within a patch to yield a 4-bit string for each patch, representing an integer from 0 to 15. This comparison mask acts as a sliding window over the image, and a 16-bin histogram is created from the full set of integers. To offer tolerance to camera translation, we extend LBP by comparing local regions rather than individual pixels, with each region the average of its underlying pixels, as shown in Fig.\ref{fig:blbp}(a).

To encode global geometry such that retargeting ensures similarity at multiple scales, we adopt the spatial pyramid pooling method \cite{Lazebnik2006} which divides an image into a set of coarse-to-fine levels. As shown in Fig.\ref{fig:blbp}(b), we perform pooling with three levels, where the second and third levels are divided into $2\times2$ and $4\times4$ partitions, respectively, with each partition assigned its own histogram based on the patches it contains. For the second and third levels, further overlapped partitions of $1\times1$ and $3\times3$ are created to allow for deformation and scale variance. For patches of $3\times3$ regions, we use patches of $24\times24$, $12\times12$ and $6\times6$ pixels for the first, second and third levels, respectively. The histograms for all partitions over all levels are then concatenated to create a $496$-\emph{d} descriptor.

\subsection{Compressing the Descriptor into a Compact Binary Code}
Recent advances in large-scale image retrieval propose compressing image descriptors into compact binary codes (known as Hashing \cite{Liu2011,Liu2012,Gong2013,Lin2015}), to allow for efficient descriptor matching in Hamming space. To enable real-time retargeting, and hence application without affecting the existing clinical workflow, we similarly compress our descriptor via a novel random forests hash. Furthermore, we propose to learn the hash function with a loss function, which maps to a new space where images from the same scene have a smaller descriptor distance, compared with the original descriptor.

Let us consider a set of training image descriptors $\left\lbrace \mathbf{x}_{i} \right\rbrace_{i=1}^{n}$ from the diagnosis sequence, each assigned to a scene label representing its topological location, where each scene is formed of a cluster of adjacent images. We now aim to infer a binary code of $m$ bits for each descriptor, by encouraging the Hamming distance between the codes of two images to be small for images of the same scene, and large for images of different scenes, as in \cite{Lin2015}. Let us now denote $\mathbf{Y}$ as an affinity matrix, where $y_{ij}=1$ if images ${x}_{i}$ and ${x}_{j}$ have the same scene label, and $y_{ij}=0$ if not. We now sequentially optimise each bit in the code, such that for $r$-th bit optimisation, we have the objective function:
\begin{equation}
\label{eq:objective}
\min_{\mathbf{b}_{\left(r\right)}} \sum_{i=1}^{n}\sum_{j=1}^{n}l_r\left( b_{r,i}, b_{r,j};y_{ij} \right), \text{   s.t. } \mathbf{b}_{\left(r\right)}\in \left\lbrace0,1 \right\rbrace^{n}.
\end{equation}
Here, $b_{r,i}$ is the $r$-th bit of image $\mathbf{x}_{i}$, $\mathbf{b}_{\left(r\right)}$ is a vector of the $r$-th bits for all $n$ images, and $l_r\left(\cdot \right)$ is the loss function for the assignment of bits $b_{r,i}$ and $b_{r,j}$ given the image affinity $y_{ij}$. As proved in \cite{Lin2015}, this objective can be optimised by formulating a quadratic hinge loss function as follows:
\begin{equation}
\label{eq:loss}
    l_r\left( b_{r,i}, b_{r,j};y_{ij}\right)=
\begin{cases}
    \left[0-\mathcal{D}\left(\mathbf{b}_{i}^r,\mathbf{b}_{j}^r \right) \right]^{2},& \text{if } y_{ij} = 1\\
    \left[ \max \left(0.5m -\mathcal{D}\left( \mathbf{b}_{i}^r, \mathbf{b}_{j}^r \right),0\right) \right]^{2},& \text{if } y_{ij} = 0
\end{cases}
\end{equation}

Here, $\mathcal{D}\left( \mathbf{b}_{i}^r, \mathbf{b}_{j}^r \right)$ denotes the Hamming distance between $\mathbf{b}_{i}$ and $\mathbf{b}_{j}$ for the first $r$ bits. Note that during binary code inference, the optimisation of each bit uses the results of the optimisation of the previous bits, and hence this is a series of local optimisations due to the intractability of global optimisation.

\subsection{Learning Encoding Functions with Random Forests}

With each training image descriptor assigned a compact binary code, we now propose a novel method for mapping $\left\lbrace \mathbf{x}_{i} \right\rbrace_{i=1}^{n}$ to $\left\lbrace \mathbf{b}_{i} \right\rbrace_{i=1}^{n}$, such that the binary code for a new query image may be computed. We denote this function $\Phi\left(\mathbf{x} \right)$, and represent it as a set of independent hashing functions $\left\lbrace \phi_{i}\left( \mathbf{x} \right) \right\rbrace_{i=1}^{m}$, one for each bit. To learn the hashing function $\phi_{i}$ of the $i^{th}$ bit in $\mathbf{b}$, we treat this as a binary classifier which is trained on input data $\left\lbrace \mathbf{x}_{i} \right\rbrace_{i=1}^{n}$ with labels $ \mathbf{b}_{\left( i \right)}$.

Rather than using boosted trees as in \cite{Lin2015}, we employ random forests \cite{Criminisi2013}, which are faster for training and less susceptible to overfitting. Our approach allows for fast hashing which enables encoding to be achieved without slowing down the clinical workflow. We create a set of random forests, one for each hashing function $\left\lbrace \phi_{i}\left( \mathbf{x} \right) \right\rbrace_{i=1}^{m}$. Each tree in one forest is independently trained with a random subset of $\left\lbrace \mathbf{x}_{i} \right\rbrace_{i=1}^{n}$, and comparisons of random pairs of descriptor elements as the split functions. We grow each binary decision tree by maximising the information gain to optimally split the data $X$ into left $X_{L}$ and right $X_{R}$ subsets at each node. This information gain $I$ is defined as:

\begin{equation}
I=\pi\left( X \right)-\dfrac{1}{|X|}\sum_{k\in\left\lbrace L,R \right\rbrace}|X_{k}|\pi\left(X_{k}\right)
\end{equation}
where $\pi\left( X \right)$ is the Shannon entropy: $
\pi\left( X \right)=-\sum_{y\in\left\lbrace 0,1 \right\rbrace}p_{y}\log\left( p_{y} \right)
$. Here, $p_{y}$ is the fraction of data in $X$ assigned to label $y$. Tree growth terminates when the tree reaches a defined maximum depth, or $I$ is below a certain threshold ($e^{-10}$). With $T$ trained trees, each returning a value $\alpha_{t}\left( \mathbf{x} \right)$ between 0 and 1, the hashing function for the $i^{th}$ bit then averages the responses from all trees and rounds this accordingly to either 0 or 1:

\begin{equation}
    \phi_{i}\left( \mathbf{x} \right)=
\begin{cases}
    0& \text{if } \frac{1}{T}\sum_{t=1}^{T} \alpha_{t}\left( \mathbf{x} \right) < 0.5\\
    1& \text{otherwise}
\end{cases}
\end{equation}

Finally, to generate the $m$-bit binary code, the mapping function $\Phi\left(\mathbf{x} \right)$ concatenates the output bits from all hashing functions $\left\lbrace \phi_{i}\left( \mathbf{x} \right) \right\rbrace_{i=1}^{m}$ into a single binary string. Therefore, to achieve retargeting, the binary string assigned to a query image from the surveillance sequence is compared, via Hamming distance, to the binary strings of scenes captured in a previous diagnosis sequence.
\section{Experiments and Results}
For validation, \emph{in vivo} experiments were performed on 13 GI videos ($\approx17,700$ images) obtained from six patients. For each from patients 1-5, two videos were recorded in two separate endoscopies of the same examination, resulting in ten videos. For patient 6, three videos were collected in three serial examinations, with each consecutive examination 3-4 months apart. All videos were collected using standard Olympus endoscopes, with NBI-mode on for image enhancement. The videos were captured at 720x576-pixels, and the black borders in the images were cropped out. 

We used leave-one-video-out cross validation, where one surveillance video ($\mathcal{O}1$) and one diagnosis video ($\mathcal{O}2$) are selected for each experiment, for a total of 16 experiments (two for each of patients 1-5, and six for patient 6). Intensity-based k-means clustering was used to divide $\mathcal{O}2$ into clusters, with the number of clusters defined empirically and proportional to the video length ($10\textup{--}34$ clusters). To assign ground truth labels to test images, videos $\mathcal{O}1$ and $\mathcal{O}2$ were observed side-by-side manually by an expert, moving simultaneously from start to end. For each experiment, we randomly selected 50 images from $\mathcal{O}1$ (testing) as queries. Our framework has been implemented using Matlab and \texttt{C++}, and runs on an HP workstation (Intel x5650 CPU).
\begin{figure*}[t]
\centering
\includegraphics[width=\linewidth]{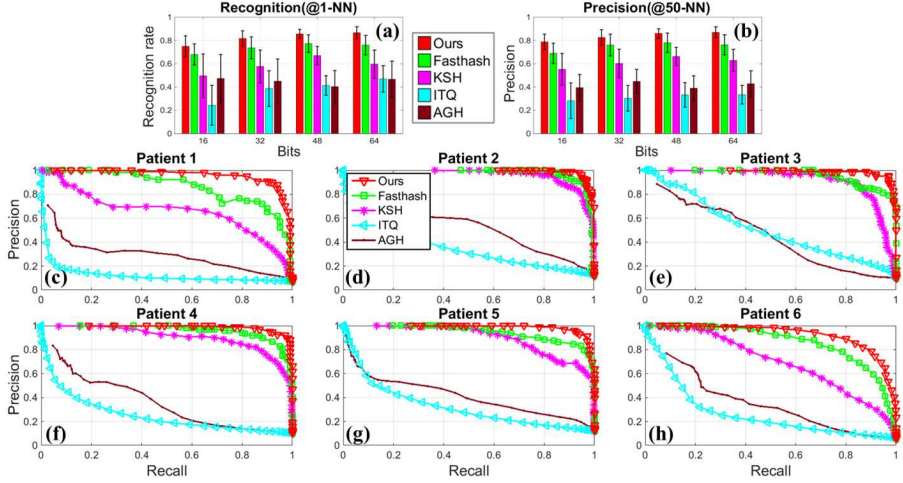}
\caption{(a) Means and standard deviations of recognition rates (precisions @ 1-NN) and (b) Precision values @ 50-NN with different binary code lengths; (c-h) Precision-recall curves of individual experiments using 64-bit codes.}
\label{fig:precrec}
\end{figure*}
\begin{table}[t]
\scriptsize
\centering
\caption{Mean average precision for recognition, both for the descriptor and the entire framework. Note that the results of hashing-based methods are at 64-bit.}
\begin{tabularx}{0.98\textwidth}{c|*{4}{Y}||*{6}{Y}}
\hline
        & \multicolumn{4}{c||}{\textbf{Descriptor}} & \multicolumn{6}{c}{\textbf{Framework}} \\ 
Methods & BOW    & GIST  & SPACT  & \textbf{Ours} & EVM & AGH  & ITQ  & KSH  & Fasthash & \textbf{Ours} \\ \hline
Pat.1   & 0.227   & 0.387  & 0.411   & \textbf{0.488} & 0.238 & 0.340 & 0.145 & 0.686 & 0.802     & \textbf{0.920} \\ 
Pat.2   & 0.307   & 0.636  & 0.477   & \textbf{0.722} & 0.304 & 0.579 & 0.408 & 0.921 & 0.925     & \textbf{0.956} \\ 
Pat.3   & 0.321   & 0.576  & 0.595   & \textbf{0.705} & 0.248 & 0.501 & 0.567 & 0.903 & 0.911     & \textbf{0.969} \\ 
Pat.4   & 0.331   & 0.495  & 0.412   & \textbf{0.573} & 0.274 & 0.388 & 0.289 & 0.889 & 0.923     & \textbf{0.957} \\ 
Pat.5   & 0.341   & 0.415  & 0.389   & \textbf{0.556} & 0.396  & 0.435 & 0.342 & 0.883 & 0.896     & \textbf{0.952} \\ 
Pat.6   & 0.201   & 0.345  & 0.315   & \textbf{0.547} & 0.273 & 0.393 & 0.298 & 0.669 & 0.812     & \textbf{0.895} \\ \hline
\end{tabularx}
\label{tab:map}
\end{table} 

Recognition results for our original descriptor before binary encoding were compared to a range of standard image descriptors, including a BOW vector \cite{Philbin2007} based on SIFT features, a global descriptor GIST based on frequency response \cite{Oliva2001}, and SPACT \cite{Wu2011}, a global descriptor based on pixel comparisons. We used the publicly-available code of GIST, and implemented a $10,000$-\emph{d} BOW descriptor and a $1,240$-\emph{d} SPACT descriptor. Descriptor similarity was computed using the L2 distance for all methods. Table \ref{tab:map} shows the mAP results, with our descriptor significantly outperforming all competitors. As expected, BOW offers poor results due to the inconsistency of local keypoint detection over long time intervals. We also outperform SPACT as the latter is based on pixel-level comparisons, while our regional comparisons are more robust to illumination variation and camera translation. Whilst GIST typically offers good tolerance to scene deformation, it lacks local texture encoding, whereas the multi-level nature of our novel descriptor ensures that similar descriptors suggest image similarity across a range of scales.

\begin{figure*}[t]
\centering
\includegraphics[width=\linewidth]{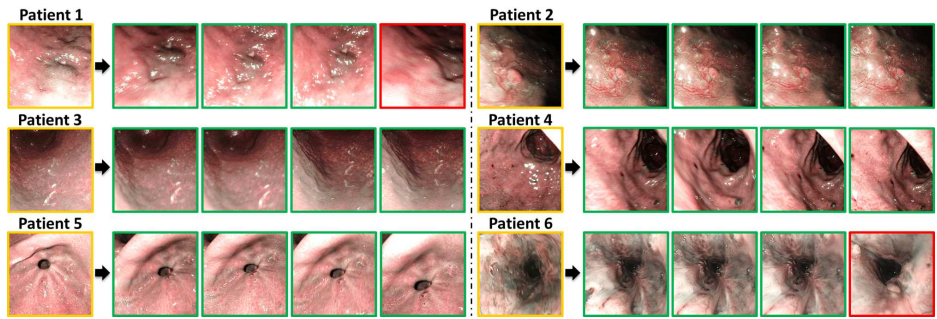}
\caption{Example top-ranked images for the proposed framework on six patients. Yellow-border images are queries from a surveillance sequence, green- and red-border images are the correctly and incorrectly matches from a diagnosis sequence, respectively.}
\label{fig:sceneres}
\end{figure*}

Our entire framework was compared to the EVM method \cite{Atasoy2012} and hashing-based methods, including ITQ\cite{Gong2013}, AGH\cite{Liu2011}, KSH\cite{Liu2012} and Fasthash\cite{Lin2015}. For the competitors based on hashing, our descriptor was used as input. For our framework, the random forest consisted of 100 trees, each with a stopping criteria of maximum tree depth of 10, or minimum information gain of $e^{-10}$. Fig.\ref{fig:precrec}(a) shows the recognition rate if the best-matched image is correct (average precision at 1-nearest-neighbour(NN)). We compare across a range of binary string lengths, with our framework consistently outperforming others and with the highest mean recognition rate $\left\lbrace0.87, 0.86, 0.82, 0.75\right\rbrace$. We also show the precision values at 50-NN in Fig.\ref{fig:precrec}(b). Precision-recall curves (at 64-bit length) for each patient data are shown in Fig.\ref{fig:precrec}(c-h), with mAP values in Table \ref{tab:map}. As well as surpassing the original descriptor, our full framework outperforms all other hashing methods, with the highest mAP scores and graceful fall-offs in precision-recall. Our separation of encoding and hashing achieves strong discrimination through a powerful independent classifier compared to the single-stage approaches of \cite{Liu2011,Liu2012,Gong2013} and the less flexible classifier of \cite{Lin2015}. We also found that the performance of EVM is inferior to ours (Table \ref{tab:map}), and significantly lower than that presented in \cite{Atasoy2012}. This is because in their work, training and testing data were from the same video sequence. In our experiments however, two different sequences were used for training and testing, yielding a more challenging task, to fully evaluate the performance on inter-examination retargeting. The current average querying time using 64-bit strings (including descriptor extraction, binary encoding and Hamming distance calculation) is around $19ms$, which demonstrates its real-time capability, compared to $490ms$ for querying with the original descriptor. Finally, images for example retargeting attempts are provided for our framework in Fig.\ref{fig:sceneres}.

Note that our descriptor currently does not explicitly address rotation invariance. However, from the experiments, we do observe the framework allows for a moderate degree of rotations. In addition, an effective way to achieving rotation invariance is to generate rotated versions of images in the diagnosis video before descriptor computation and hashing. 
\section{Conclusions}
In this paper, we have proposed a retargeting framework for optical biopsy in serial endoscopic examinations. A novel global image descriptor with regional comparisons over multiple scales deals with tissue appearance variation across examinations, whilst binary encoding with a novel random forest-based mapping function adds discrimination and speeds up recognition. The framework can be readily incorporated into the existing endoscopic workflow due to its capability of real-time retargeting and no requirement of manual calibration. Validation on \emph{in vivo} videos of serial endoscopies from six patients, shows that both our descriptor and hashing scheme are consistently state-of-the-art.

\bibliographystyle{splncs03}
\bibliography{miccai}
\end{document}